%% file: final.tex
\documentclass{article}

\usepackage{microtype}
\usepackage{graphicx}
\usepackage{subfigure}
\usepackage{booktabs} % for professional tables
\usepackage{bm}
\usepackage{bbm}
\usepackage{amsmath,amsfonts,amssymb,amsthm}
\usepackage{mathtools}

\input{definitions.tex}

\usepackage{hyperref}

\DeclarePairedDelimiter\ceil{\lceil}{\rceil}
\DeclareMathOperator*{\E}{\mathbb{E}}
\DeclareMathOperator*{\R}{\mathbb{R}}
\DeclareMathOperator*{\f}{f(\xn \,|\, \theta)}
\DeclareMathOperator*{\fz}{f(\bm \ast \,|\, \theta)}
\DeclareMathOperator*{\fdeep}{g(\xn \,|\, \theta)}
\newcommand{\wa}{w^{[0]}_{\gamma}}
\newcommand{\wb}{w^{[1]}_{k}}
\newcommand{\wt}{w^{(t)}_{k}}

\newcommand{\wtt}{w^{(t+1)}_{k}}
\newcommand{\wat}{w^{[0](t)}_{\gamma}}
\newcommand{\wbt}{w^{[1](t)}_{k}}

\newcommand{\watt}{w^{[0](t+1)}_{\gamma}}
\newcommand{\wbtt}{w^{[1](t+1)}_{k}}

\newtheorem{observation}{Observation}
\newtheorem{claim}{Claim}
\newtheorem{definition}{Definition}

\usepackage{tikz}
\usepackage{standalone}
\usetikzlibrary{shapes.geometric}
\usetikzlibrary{arrows}
\usetikzlibrary{arrows.meta}

\usepackage{xcolor}

\usepackage[accepted]{icml2019}

\icmltitlerunning{Optimisation of Overparametrized Sum-Product Networks}

\begin{document}

\twocolumn[
\icmltitle{Optimisation of Overparametrized Sum-Product Networks}

\icmlsetsymbol{equal}{*}

\begin{icmlauthorlist}
\icmlauthor{Martin Trapp}{ofai,graz}
\icmlauthor{Robert Peharz}{cam}
\icmlauthor{Franz Pernkopf}{graz}
\end{icmlauthorlist}

\icmlaffiliation{ofai}{Austrian Research Institute for AI, Austria}
\icmlaffiliation{cam}{CBL Lab, Department of Engineering, University of Cambridge, United Kingdom}
\icmlaffiliation{graz}{SPSC Lab, Graz University of Technology, Austria}

\icmlcorrespondingauthor{Martin Trapp}{martin.trapp@tugraz.at}

\icmlkeywords{Sum-Product Networks, Gradient Optimisation}

\vskip 0.3in
]

\printAffiliationsAndNotice{}

\begin{abstract}
It seems to be a pearl of conventional wisdom that parameter learning in deep sum-product networks is surprisingly fast compared to shallow mixture models.
This paper examines the effects of overparameterization in sum-product networks on the speed of parameter optimisation.
Using theoretical analysis and empirical experiments, we show that deep sum-product networks exhibit an implicit acceleration compared to their shallow counterpart.
In fact, gradient-based optimisation in deep tree-structured sum-product networks is equal to gradient ascend with adaptive and time-varying learning rates and additional momentum terms.
\end{abstract}

\section{Introduction}
\label{introduction}
Tractable probabilistic models, in which exact inference can be tackled efficiently, have gained increasing popularity within and outside the machine learning community.
In particular Sum-Product Networks (SPNs) \cite{Poon2011a}, - which subsume existing approaches such as latent tree models and deep mixture models \cite{Jaini2018} - have shown to perform well on various tasks, e.g. image classification \cite{Gens2012, Peharz2018}, action recognition \cite{Amer2016}, bandwidth extension \cite{Peharz2014}, language modelling \cite{Cheng2014}, spatial modelling \cite{Pronobis2017} and non-linear regression \cite{Trapp2018}.
Much of their success is due to their flexibility, tractability and efficient representation of complex function approximations.

Thus in recent years various ways to perform parameter learning in SPNs and build their structure, e.g. \cite{Gens2013,Vergari2015}, have been proposed.
For example, using a latent variable interpretation of SPNs \cite{Poon2011a,Peharz2017} one can derive classic expectation-maximization as introduced in \cite{Peharz2017}.
Moreover, parameter learning can also be tackled within the Bayesian framework using, e.g. variational inference \cite{Zhao2016} or Bayesian moment-matching \cite{Rashwan2016}.
Optimising a discriminative objective can be achieved using gradient-based optimisation \cite{Gens2012} and recently Peharz {\it et~al.} \cite{Peharz2018} introduced a hybrid objective also optimised using gradient-based optimisation.
Semi-supervised learning can be tacked likewise, as shown by Trapp {\it et~al.} \cite{Trapp2017}.
However, even though many approaches utilise gradient-based optimisation for parameter learning it is not clear if and to which extend the depth of an SPNs has an effect on the speed of optimisation.

On the other hand, analysing the dynamics of optimisation for linear neural networks, see \cite{Baldi1995} for a survey on linear neural networks, has recently gained increasing interest, e.g. \cite{Saxe2014,Arora2018}.
In particular, Arora {\it et~al.} \cite{Arora2018} have shown that increasing the depth in linear neural networks can speed up the optimisation. In fact, they showed that the acceleration effect of overparameterization in linear neural networks cannot be achieved by \emph{any} regulariser.

This work discusses the implicit acceleration effects of depth and overparameterization in SPNs, which are a multi-linear function in their weights. As SPNs have been used for non-linear classification tasks, e.g. \cite{Gens2012,Trapp2017,Peharz2018}, implicit acceleration effects due to their depth are of particular relevance.

The remainder of this paper is structured as follows:
In Section \ref{sec:background} we review recent work on overparameterization in linear neural networks and briefly introduce SPNs.
We discuss acceleration effect in SPNs in Section~\ref{sec:main} and conclude the work in Section~\ref{sec:conclusion}.

\section{Background and Related Work} \label{sec:background}
\subsection{Overparameterization in Linear Networks}
Recent work has shown that increasing depth in linear neural networks can speed up the optimisation \cite{Arora2018}. 
In fact, even gradient-based optimisation of linear regression benefits by moving from a convex objective function to a non-convex objective.
In particular, let
\[
\ell_p(\bm w) = \E_{(\bm{x},y) \sim \mathcal{D}} \left[ \frac{1}{p} (\bm x^{\top} \bm w - y)^p \right] \, ,
\]
be the $\ell_p$ loss function of a linear regression model parametrised with $\bm w \in \R^D$ and let $\bm{x} \in \R^D$ be the explanatory variables and $y \in \R$ their respective dependent variable.
Note that we use bold font to indicate vectors.
Suppose that we now artificially overparametrized this linear model as follows:
\[
\bm w = \bm w_1 \cdot w_2 \, ,
\]
for which $\bm w_1 \in \R^D$ and $w_2 \in \R$.
Thus, our overparameterization results in the following non-convex loss function:
\[
\ell_p(\bm w_1, w_2) = \E_{(\bm{x},y) \sim \mathcal{D}} \left[ \frac{1}{p} (\bm x^{\top} \bm w_1 w_2 - y)^p \right] \, .
\]

Now let us consider the respective gradients, i.e. 
\begin{align*}
  \nabla_{\bm w} &:= \E_{(\bm{x},y) \sim \mathcal{D}} \left[ (\bm x^{\top} \bm w - y)^{p-1} \bm x \right] \\
  \nabla_{\bm w_1} &:= \E_{(\bm{x},y) \sim \mathcal{D}} \left[ (\bm x^{\top} \bm w_1 w_2 - y)^{p-1} w_2 \bm x \right] \\
  \nabla_{w_2} &:= \E_{(\bm{x},y) \sim \mathcal{D}} \left[ (\bm x^{\top} \bm w_1 w_2 - y)^{p-1} \bm w_1^{\top} \bm x \right]
\end{align*}
allowing to compute the updated parameters at time $t+1$, e.g.
\[
\bm w^{(t+1)}_1 \leftarrow \bm w^{(t)}_1 - \eta \nabla_{\bm w^{(t)}_1} \, ,
\]
where $\eta$ is a fixed learning rate.

To understand the dynamics of the underlying parameter $\bm w = \bm w_1 w_2$, Arora {\it et~al.} \cite{Arora2018} show that
\begin{align*}
  \bm w^{(t+1)} &= \bm w_1^{(t+1)} \cdot w_2^{(t+1)} \\
   &= \bm w^{(t)} - \eta (w^{(t)}_2)^2 \nabla_{\bm w^{(t)}} - \eta (w^{(t)}_2)^{-1} \nabla_{w^{(t)}_2} \bm w^{(t)} \\
   &= \bm w^{(t)} - \rho^{(t)} \nabla_{\bm w^{(t)}} - \lambda^{(t)} \bm w^{(t)} \, .
\end{align*}
Initialising all weights close to zero allows arriving at an update rule that is similar to momentum optimisation \cite{Nesterov1983} with an adaptive learning rate.
Arora {\it et~al.} further show that overparameterizing fully connected linear neural networks, i.e. increasing the number of layers, leads to an adaptive learning rate and gradient projection amplification that can be thought of as momentum optimisation.

\subsection{Sum-Product Networks}
A Sum-Product Network (SPN) \cite{Poon2011a} is a rooted directed acyclic graph composed of sum nodes ($\SumNode$), product nodes ($\ProductNode$) and leaf nodes ($\Leaf$), i.e.  indicators or arbitrary distributions \cite{Peharz2015}.
Let $\{X_{d}\}_{d=1}^D$ be a set of random variables (RVs), then an SPN represents the joint $p(X_1, \dots, X_D)$ using a recursive structure in which internal nodes ($\Node$) either compute a weighted sum, i.e. $\SumNode(\x) = \sum_{\Node \in \ch(\SumNode)} w_{\SumNode,\Node} \, \Node(\x)$, 
or compute a product of the values of their children, i.e. $\ProductNode(\x) = \prod_{\Node \in \ch(\ProductNode)} \Node(\x)$,
where $\ch(\Node)$ denotes the children of node $\Node$.
Each edge $(\SumNode,\Node)$ emanating from a sum node $\SumNode$ has a non-negative weight $w_{\SumNode,\Node}$. The sum node is said to be normalised if the weights sum to one.

In the context of this work, we require SPNs to be \emph{complete} and \emph{decomposable} \cite{Poon2011a,Darwiche2003}.
Both concepts can be expressed in terms of the scope of the nodes.
The scope of a node in an SPN is the set of variables that appear in it, i.e. the node models a joint over those RV's.
An SPN is \emph{complete} if for each sum node $\SumNode$ the scopes of all children of $\SumNode$ are equal.
Further, an SPN is \emph{decomposable}, if all children of a product node have non-overlapping scopes.
The scope of each internal node is defined as the union of its children's scopes.
Due to those two conditions, SPNs allow for efficient marginalisation, conditioning, sampling, and exact inference.

\section{Overparameterization in Sum-Product Networks} \label{sec:main}
\begin{figure}[t]
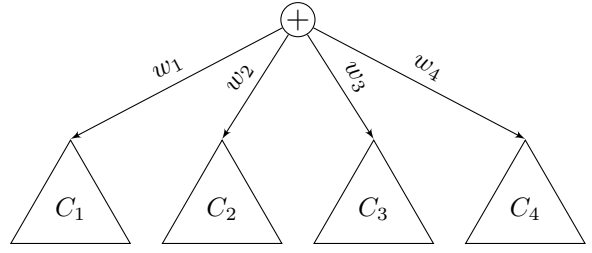

  \centering
  \includestandalone[width=\linewidth]{shallowSPN}
  \caption{Illustration of a shallow SPN. Triangles denote sub-SPNs or distributions under the root of the SPN.}
  \label{fig:shallow}
\end{figure}

For SPNs there seems to be a pearl of conventional wisdom that parameter learning is surprisingly fast.
Following the approach by Arora {\it et~al.} \cite{Arora2018} we will now discuss the implicit dynamics of gradient-based optimisation in SPNs.
To simplify the discussion, consider the network structure illustrated in Figure~\ref{fig:shallow}, i.e. an SPN with a root node that computes a weighted sum over sub-networks $C_1, \dots, C_K$.

We can denote the log-likelihood function of this network as
\begin{equation} \label{eq:shallowLLH}
\mathcal{L}(\theta\, | \, \mathcal{X}) = \sum_{n=1}^N \log \f - \log \fz \, ,
\end{equation}
where $\mathcal{X} = \{\xn\}_{n=1}^N$ is a set of observed data and $\theta$ contains all parameters of our model. Further we have
\[
\f = \sum_{k=1}^K w_k C_k(\xn) \, ,
\]
where $C_k(\bm x)$ denotes the value of child $k$ for observation $\xn$.
Note that $\fz$ is the partition function\footnote{The partition function is obtained by setting all indicators to one.}, see \cite{Poon2011a} for details. The normalisation is only necessary if the network is not normalised, i.e.  $\fz \not = 1$.
For deriving our results we will assume that all $w_k$ are initialised as $w_k \approx 0$, i.e. the network is not normalised.

To maximise Equation~\ref{eq:shallowLLH} we can use vanilla gradient ascend with the following gradients:
\begin{equation}
  \nabla_{w_k} := \frac{1}{\f}C_k(\bm x) - \frac{1}{\fz}C_k(\ast) \, ,
\end{equation}
where $C_k(\ast)$ denotes the partition function of sub-network $C_k$.
Thus optimising Equation~\ref{eq:shallowLLH} can be achieved by updating $w_k$ accordingly at each iteration $t$.

\begin{figure}[t]
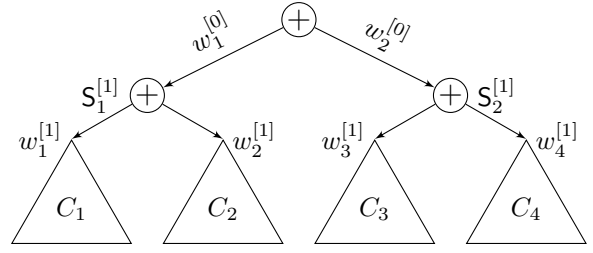

  \centering
  \includestandalone[width=\linewidth]{deepSPN}
  \caption{Illustration of an overparametrized SPN. Triangles denote sub-SPNs or distributions.}
  \label{fig:deep}
\end{figure}

Let us now consider an overparametrized version of this network by introducing additional sum nodes into the model.
For this purpose, let each weight $w_k$ be decomposed into multiple independent weights.
See Figure~\ref{fig:deep} for an illustration of an overparametrized version of our initial network.
Note that this SPN is in its expressiveness equivalent to our initial model and only introduces additional parameters.
Therefore, let $w^{[l]}_{j}$ denote the $j$\textsuperscript{th} weight of layer $l$.
We can define the decomposition of $w_k$ by $L$ sum node layers as
\begin{equation} \label{eq:decomposition}
w_k = \prod_{l=0}^L w^{[l]}_{\phi(k, l)} \, ,
\end{equation}
where $\phi(k, l)$ maps the index $k$ onto the respective index at layer $l$.
For example, in the case of our model in Figure~\ref{fig:deep} we can define $\phi(k, l)$ as
\[
    \phi(k, l) =
\begin{cases}
    \ceil*{\frac{k}{2}} ,& \text{if } l = 0\\
    k, & \text{otherwise}
\end{cases} \, ,
\]
where $\ceil*{\,}$ denotes the ceiling operator.
Note that the definition of $\phi(k, l)$ generally depends on the network structure of the SPN.

Similarly, we can denote the log-likelihood function of an overparametrized network as
\begin{equation} \label{eq:deepLLH}
\mathcal{L}(\theta\, | \, \mathcal{X}) = \sum_{n=1}^N \log \fdeep - \log \fz \, ,
\end{equation}
where
\[
\fdeep = \sum_{k=1}^K \prod_{l=0}^L w^{[l]}_{\phi(k, l)} C_k(\xn) \, .
\]
Note that $\f = \fdeep$ as we defined $w_k = \prod_{l=0}^L w^{[l]}_{\phi(k, l)}$.
Let $\gamma := \ceil*{\frac{k}{2}}$ and let us assume the SPN illustrated in Figure~\ref{fig:deep}.
Therefore, we have the decomposition 
\[
w_k = \wa \cdot \wb
\]
for each weight $k$. Thus we define the gradients of $\wa$ and $\wb$ as
\begin{align}
  \begin{aligned}
  \nabla_{\wb} &:= \frac{\wa}{\f} C_k(\xn) - \frac{\wa}{\fz} C_k(\ast) \\
  &= \wa \left[ \frac{1}{\f} C_k(\xn) - \frac{1}{\fz} C_k(\ast) \right] \\
  &= \wa \nabla_{w_k} \, ,
  \end{aligned}
\end{align}
and
\begin{align}
  \begin{aligned}
  \nabla_{\wa} &:= \sum_{j \in \ch(S^{[1]}_{\phi(k,0)})} \frac{w^{[1]}_j}{\f} C_j(\xn) - \frac{w^{[1]}_j}{\fz} C_j(\ast) \\
  &= \sum_{j \in \ch(S^{[1]}_{\phi(k,0)})} w^{[1]}_j \nabla_{w_j} \, ,
\end{aligned}
\end{align}
where $S^{[1]}_{\phi(k,0)}$ denotes the sum node at the first layer which is connected to the edge with weight $\wa$.
In more general terms we can say that the gradient of $w^{[l]}_{\phi(k, l)}$ at layer $l$ is defined by:
\begin{equation} \label{eq:generalGradients}
  \nabla w^{[l]}_{\phi(k, l)} := \sum_{w^{[l]}_{\phi(k, l)} \leadsto j} (w^{[l]}_{\phi(k, l)})^{-1} w_j \nabla{w_j} \,
\end{equation}
where we use $w^{[l]}_{\phi(k, l)} \leadsto j$ to denote the set of all weights $w_j:= \prod_{l=0}^L w^{[l]}_{\phi(j, l)}$ for which $w^{[l]}_{\phi(k, l)}$ is included in the decomposition of $w_j$.

Similar to Arora {\it et~al.} \cite{Arora2018}, we can now examine the dynamics of $w_k$ by assuming a small learning rate $\eta$ and by assuming that all weights are initialised near zero.
For this let $\wtt$ denote $w_k$ at time $t+1$. Therefore, we obtain
\begin{align} \label{eq:toyUpdateRule}
  \wtt &= \watt \cdot \wbtt \\
  &= \Bigg[ \wat + \eta \nabla_{\wat} \Bigg] \Bigg[ \wbt + \eta \nabla_{\wbt} \Bigg] \\
\begin{split}
  &= \color{red} \wat \wbt \color{black} + \color{cyan} \eta \wat \nabla_{\wbt}  \\
  &+ \color{blue} \eta \wbt \nabla_{\wat} \color{black} + \mathcal{O}(\eta^2) \, ,
\end{split}\label{eq:gradwt}
\end{align}
where we can drop $\mathcal{O}(\eta^2)$ as we assume $\eta$ to be small.
Thus we can reformulate Equation~\ref{eq:gradwt} as follows:
\begin{align}
\begin{split}
  \wtt &\approx \color{red} \wt \color{black} + \color{cyan} \eta (\wat)^{2}\nabla_{\wt} \\
  &+ \color{blue} \eta \nabla_{\wat} (\wat)^{-1} \wt
\end{split} \\
  &= \color{red} \wt \color{black} + \color{cyan} \rho^{(t)} \nabla_{\wt} \color{black} + \color{blue} \lambda^{(t)} \wt \,\color{black} .
\end{align}
We can find the solution for any decomposition, c.f. Equation~\ref{eq:decomposition},  by considering the gradients in Equation~\ref{eq:generalGradients}. 
Thus the update rule of the weights for any overparametrized SPNs is
\begin{align}\label{eq:generalUpdateRule}
  \begin{aligned}
  \wtt &\approx \wt + \eta (w^{[0]}_{\phi(k, 0)})^{2}\nabla_{\wt} \\
  &+ \left[ \sum_{l=0}^{L-1} \eta \nabla_{w^{[l]}_{\phi(k, l)}} (w^{[l]}_{\phi(k, l)})^{-1} \right] \wt \, ,
  \end{aligned}
\end{align}
where we drop terms in which we have $\eta$ to the power of two or larger as we assume $\eta$ to be small.
We can now define the adaptive and time-varying learning rate
\[
\rho^{(t)} := \eta (w^{[0]}_{\phi(k, 0)})^{2} \, ,
\]
and let
\[
\lambda^{(t)} := \sum_{l=0}^{L-1} \eta \nabla_{w^{[l]}_{\phi(k, l)}} (w^{[l]}_{\phi(k, l)})^{-1} \, .
\]
Note that we pulled out $l=0$ to obtain the gradient of $w_k$, c.f. second term in Equation~\ref{eq:generalUpdateRule}, thus resulting in the weight $\lambda^{(t)}$ to be a summation over $L-1$ terms.
Therefore, we can see that the gradient updates of $w_k$ are directly influenced by the depth of the network.

Let all weights be initialisation near zero, then $\wt$ can be understood as a weighted combination of past gradients and thus there exists a $\mu^{(t,\tau)} \in \R$ such that the dynamics of $\wt$ correspond to gradient optimisation with momentum, i.e. 
\begin{equation}
  \wt \approx \wt + \rho^{(t)} \nabla_{\wt} + \sum_{\tau=1}^{t-1} \mu^{(t,\tau)} \nabla_{w^{(\tau)}_k} \, .
\end{equation}

\begin{observation}
Gradient-based optimisation of an overparametrized sum-product network with small (fixed) learning rate and near zero initialisation of the weights is equivalent to gradient-based optimisation with adaptive and time-varying learning rate and momentum terms.
\end{observation}

So far we have only considered the case in which we artificially introduce additional sum nodes to the network.
However, an important question is if the depth of an SPN also implicitly accelerates parameter learning.
As shown by \cite{Zhao2016} the density function of an SPN can be written as a mixture over induced trees, i.e. 
\[
\f = \sum_{k=1}^{\kappa} \prod_{w_{S,C} \in \mathcal{T}_k} w_{S,C} \prod_{\Leaf \in \mathcal{T}_k} p(\xn \,|\, \theta_{\Leaf}) \, ,
\]
where $\mathcal{T}_k$ denotes the $k$\textsuperscript{th} induced tree, $\kappa$ is the number of induced trees and $\theta_{\Leaf}$ indicates the parameter of the leaf $\Leaf$ which is included in $\mathcal{T}_k$.

\begin{definition}[Induced tree]
Let $\SPN$ be a complete and decomposable (SPN).
An induced tree $\SPT$ of $\SPN$ is a sub-tree of $\SPN$ for which i) the root of $\SPT$ is the root of $\SPN$, ii) each sum node $\SumNode$ in $\SPT$ has only one child $\Child$ and both $\SumNode$ and $\Child$ as well as $w_{\SumNode,\Child}$ are in $\SPN$, iii) each product node $\ProductNode$ in $\SPT$ has the same children as in $\SPN$.
\end{definition}

Given a representation of a tree-structured SPN by a mixture of $\kappa$ many induced trees, for every component $k$ the weight decomposition of its component is given by the respective induced tree, i.e.
\[
  w_k = \prod_{w_{S,C} \in \mathcal{T}_k} w_{S,C} \, .
\]
Therefore, there exists an index function $\phi: \mathbb{Z} \times \mathbb{Z} \rightarrow \SumNodes \times \Nodes$, where $\SumNodes$ is the set of sum nodes and $\Nodes$ the set of nodes, such that the decompositions of the components weights can be represented by an overparametrized SPN without changing the decompositions, the order of the weights or the weights itself, i.e.
\[
\prod_{l=0}^{L_{\mathcal{T}_k}} w^{[l]}_{\phi(k, l)} := \prod_{w_{S,C} \in \mathcal{T}_k} w_{S,C} \, 
\]
where $L_{\mathcal{T}_k}$ is the depth of the induced tree.

Thus, we can say that gradient-based optimisation using $\prod_{w_{S,C} \in \mathcal{T}_k} w_{S,C}$ is equivalent to gradient-based optimisation using $\prod_{l=0}^{L_{\mathcal{T}_k}} w^{[l]}_{\phi(k, l)}$ and entails the same acceleration effects. 

\begin{claim}
Gradient-based optimisation of any deep tree-structured sum-product network with small (fixed) learning rate and near zero initialisation of the weights is equivalent to gradient-based optimisation with adaptive and time-varying learning rate and momentum terms.
\end{claim}

Note that it is, therefore, not necessary to explicitly obtain an overparametrized SPN from a naturally deep SPN.
Further, the same result can be shown for supervised and semi-supervised learning of SPNs with linear or non-linear leaves, e.g. as used in \cite{Gens2012,Trapp2017,Peharz2018}.
Thus, indicating that the depth of a network can help to accelerate parameter optimisation of non-linear classifiers if the non-linearities only occur at the terminal nodes.

\subsection{Empirical Results}
\begin{figure}[t]
  \includegraphics[width=\linewidth]{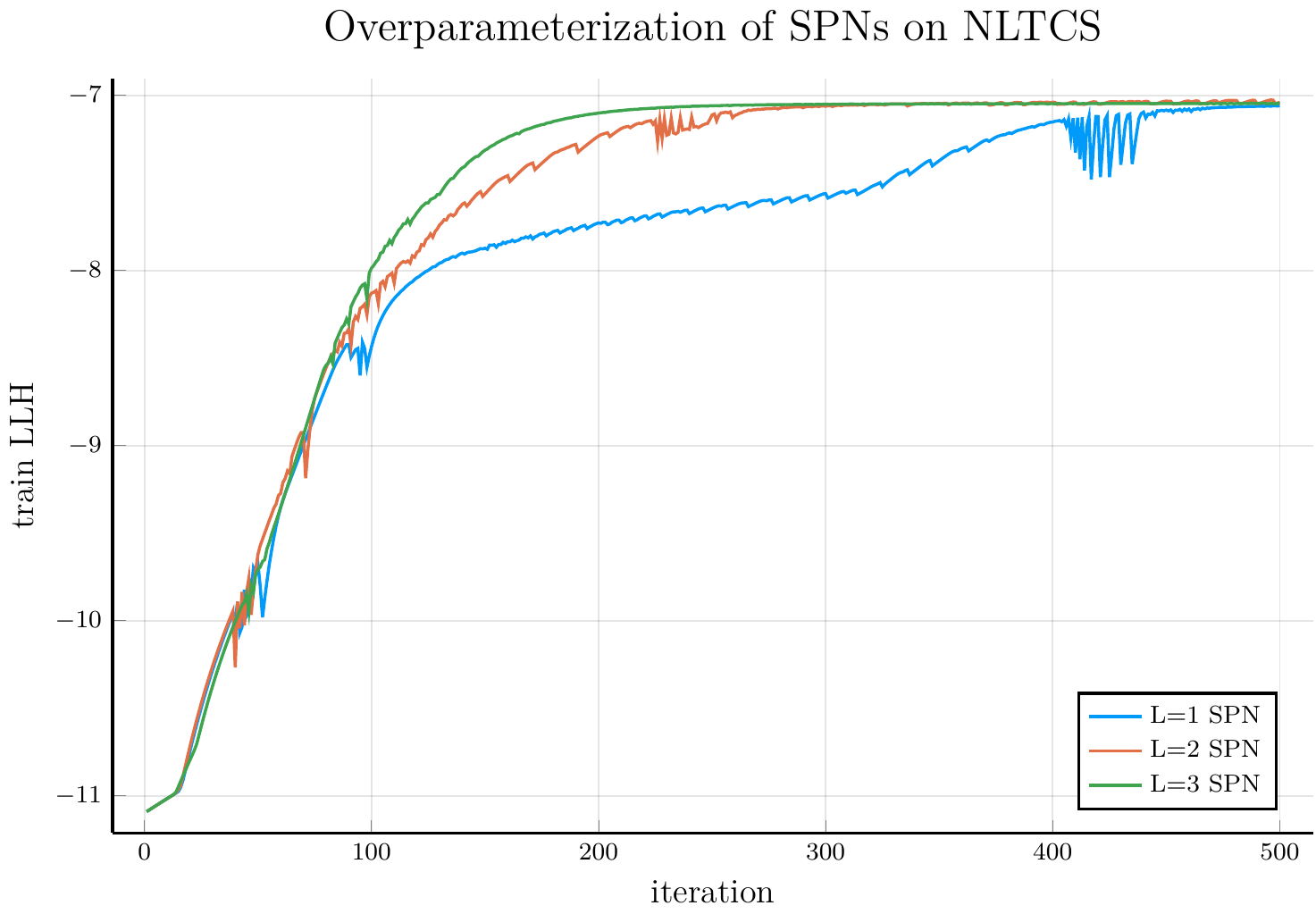}
  \caption{Empirical evaluation of overparameterization in sum-product networks (SPNs). $L$ denotes the number of consecutive sum layers. We can see that adding additional parameters helps gradient-based optimisation and leads to an implicitly acceleration, higher LLH values are better. (Best seen in colour.)}
  \label{fig:experiment}
\end{figure}

To empirically evaluate the effects of overparameterization in SPNs, we compared the speed of optimisation of a shallow SPN (similar to the one shown in Figure~\ref{fig:shallow}) to deep variants with the same expressiveness as the shallow model.\footnote{The code to reproduce the experiment can be found on GitHub under \url{https://github.com/trappmartin/TPM2019}.}
We used SPNs with 8 components ($C_1, \dots, C_8$) and initialised all weights close to zero.
Further we used vanilla gradient ascend for $500$ iterations with a learning rate of $\eta = 0.01$.
We used data extracted from the National Long Term Care Survey (NLTCS)\footnote{\url{http://lib.stat.cmu.edu/datasets/}} to examine the effects of overparameterization which contains 16 binary variables representing functional disability measures. 
We used only the training set, as processed by \cite{Lowd2010}, consisting of 16.181 observed samples to train each model.
For each iteration we reported the log-likelihood on the training set (train LLH) resulting in the curves shown in Figure~\ref{fig:experiment}.
Note that we estimated the train LLH curves over 2 independent re-runs to account for random effects of the structure generation.

We can see from the empirical experiment that increasing the number of sum layers, which is equivalent to increasing the depth of the network, leads to faster parameter optimisation.
In fact, using $L=3$ consecutive sum layers accelerates the optimisation in a way that a near optimal solution is found after only half of the number of iterations compared to the shallow construction, i.e. $L=1$.

\section{Conclusion} \label{sec:conclusion}
We have shown that overparameterization of sum-product networks (SPNs) has similar dynamics as observed in linear neural networks.
In fact, we observe that the dynamics in SPNs correspond to gradient optimisation with adaptive and time-varying learning rate and momentum terms leading to an implicit acceleration of gradient-based optimisation.
Further, we showed that naturally deep tree-structured SPNs entail the same acceleration effects as overparametrized SPNs. 
The acceleration effects in SPNs are not completely surprising as SPNs are multi-linear functions in their weights which can be represented by sparsely connected linear-neural networks with potentially non-linear input units.
As SPNs have been used for non-linear classification/modelling tasks frequently in recent years, the observation that the depth of such networks accelerates parameter learning is of particular relevance.

\section*{Acknowledgements}
This research was partially funded by the Austrian Science Fund (FWF) under the project number I2706-N31.

\bibliography{literatur}
\bibliographystyle{icml2019}

\end{document}

%% file: definitions.tex
\newcommand{\SPT}{\mathcal{T}}

\newcommand{\SPN}{\mathcal{S}}
\newcommand{\ProductNode}{\mathsf{P}}

\newcommand{\SumNode}{\mathsf{S}}
\newcommand{\SumNodes}{\bm{\mathsf{S}}}
\newcommand{\Leaf}{\mathsf{L}}

\newcommand{\Node}{\mathsf{N}}
\newcommand{\Nodes}{\bm{\mathsf{N}}}
\newcommand{\Child}{\mathsf{C}}

\newcommand{\x}{\mathbf{x}}
\newcommand{\xn}{\mathbf{x}_{n}}

\newcommand{\ch}{\ensuremath{\mathbf{ch}}}

%% file: shallowSPN.tex
  
\tikzstyle{vertex}=[inner sep=0.01cm, circle, draw]
\resizebox{!}{0.8\textheight}{
  \begin{tikzpicture} [
  scale=1, 
  auto, >={latex'[width=5cm]},
  triangle/.style = {regular polygon, regular polygon sides=3, draw },
]
    \path[use as bounding box] (-4,0.5) rectangle (4,-3);
    
    \node[vertex](root) at (0, 0) {\large$+$};
    
    \node[triangle](j1) at (-3, -2.5) {$C_1$};
    \node[triangle](j2) at (-1, -2.5) {$C_2$};
    \node[triangle](j3) at (1, -2.5) {$C_3$};
    \node[triangle](j4) at (3, -2.5) {$C_4$};
    
    \draw[->] (root) -- (j1.north) node[above, midway, sloped] {$w_1$};
    \draw[->] (root) -- (j2.north) node[above, midway, sloped] {$w_2$};
    \draw[->] (root) -- (j3.north) node[above, midway, sloped] {$w_3$};
    \draw[->] (root) -- (j4.north) node[above, midway, sloped] {$w_4$};
	
 	\end{tikzpicture}
 	}

%% file: deepSPN.tex
  
\tikzstyle{vertex}=[inner sep=0.01cm, circle, draw]
\resizebox{!}{0.8\textheight}{
  \begin{tikzpicture} [
  auto, >={latex'[width=5cm]},
  triangle/.style = {regular polygon, regular polygon sides=3, draw },
]
    \path[use as bounding box] (-4,0.5) rectangle (4,-3);
    
    \node[vertex](root) at (0, 0) {\large$+$};
    
    \node[vertex, label=left:$\mathsf{S}^{[1]}_1$](s1) at (-2, -1.0) {\large$+$};
    \node[vertex, label=right:$\mathsf{S}^{[1]}_2$](s2) at (2, -1.0) {\large$+$};
    
    \node[triangle](j1) at (-3, -2.5) {$C_1$};
    \node[triangle](j2) at (-1, -2.5) {$C_2$};
    \node[triangle](j3) at (1, -2.5) {$C_3$};
    \node[triangle](j4) at (3, -2.5) {$C_4$};
    
    \draw[->] (root) -- (s1) node[above, midway, sloped] {$w^{[0]}_1$};
    \draw[->] (root) -- (s2) node[above, midway, sloped] {$w^{[0]}_2$};
    
    \draw[->] (s1) -- (j1.north) node [anchor=east] {$w^{[1]}_1$};
    \draw[->] (s1) -- (j2.north) node [anchor=west] {$w^{[1]}_2$};
    \draw[->] (s2) -- (j3.north) node [anchor=east] {$w^{[1]}_3$};
    \draw[->] (s2) -- (j4.north) node [anchor=west] {$w^{[1]}_4$};
	
 	\end{tikzpicture}
 	}